\documentclass[10pt,twocolumn,letterpaper]{article}

\usepackage[pagenumbers]{wacv} 

\definecolor{wacvblue}{rgb}{0.21,0.49,0.74}
\usepackage[pagebackref,breaklinks,colorlinks,allcolors=wacvblue]{hyperref}
\usepackage[table]{xcolor}
\usepackage{xcolor} 
\usepackage[normalem]{ulem}
\usepackage{amsmath}

\def\wacvPaperID{1985} 
\def\confName{WACV}
\def\confYear{2026}

\title{6D Strawberry Pose Estimation: Real-time and Edge AI Solutions Using Purely Synthetic
Training Data}

\author{
Saptarshi Neil Sinha\\
Fraunhofer IGD\\
{\tt\small saptarshi.neil.sinha@igd.fraunhofer.de}
\and
Paul Julius Kühn\\
Fraunhofer IGD\\
{\tt\small julius.kuehn@igd.fraunhofer.de}
\and
Mika Silvan Goschke \\
Fraunhofer IGD\\
{\tt\small mika.silvan.goschke@igd.fraunhofer.de}
\and 
Michael Weinmann \\
Delft University of Technology\\
{\tt\small m.weinmann@tudelft.nl}
}

\begin{document}
\def\ie{i.e.\ }
\def\eg{e.g.\ }
\def\etal{et~al.\ }
\def\wrt{w.r.t.\ }


\newcommand{\todo}[1]{\textcolor{red}{#1}}
\newcommand{\mwrevision}[2]{{\color{red}\sout{#1}}{\color{green}\uwave{#2}}}
\newcommand{\nrevision}[2]{{\color{red}\sout{#1}}{\color{blue}\uwave{#2}}}
\newcommand{\finalrevision}[2]{{}{\color{black} #2}}
\maketitle
\begin{abstract}
Automated and selective harvesting of fruits has become an important area of research, particularly due to challenges such as high costs and a shortage of seasonal labor in advanced economies. This paper focuses on 6D pose estimation of strawberries using \finalrevision{}{purely} synthetic data generated through a procedural pipeline for photorealistic rendering. We employ the YOLOX-6D-Pose algorithm, a single-shot approach that leverages the YOLOX backbone, known for its balance between speed and accuracy, and its support for edge inference.
 \finalrevision{}{To address the lacking availability of training data}, we introduce a robust and flexible pipeline for generating synthetic strawberry data from various 3D models via a procedural Blender pipeline, \finalrevision{}{ where we focus on enhancing the realism of the synthesized data in comparison to previous work to make it a valuable resource for training pose estimation algorithms}. Quantitative evaluations indicate that \finalrevision{}{our models achieve comparable accuracy on both the NVIDIA RTX 3090
and Jetson Orin Nano} across several ADD-S metrics, with the RTX 3090 demonstrating superior processing speed. However, the Jetson Orin Nano is particularly suited for resource-constrained environments, making it an excellent choice for deployment in agricultural robotics.
Qualitative assessments further confirm the model's performance, demonstrating its capability to accurately infer the poses of ripe and partially ripe strawberries, while facing challenges in detecting unripe specimens. This suggests opportunities for future improvements, especially in enhancing detection capabilities for unripe strawberries (if desired) by exploring variations in color. Furthermore, the methodology presented could be adapted easily for other fruits such as apples, peaches, and plums, thereby expanding its applicability and impact in the field of agricultural automation.
\end{abstract}

\section{Introduction}
\sloppy
\begin{figure*}[htb!]
    \centering
    \includegraphics[width=\linewidth]{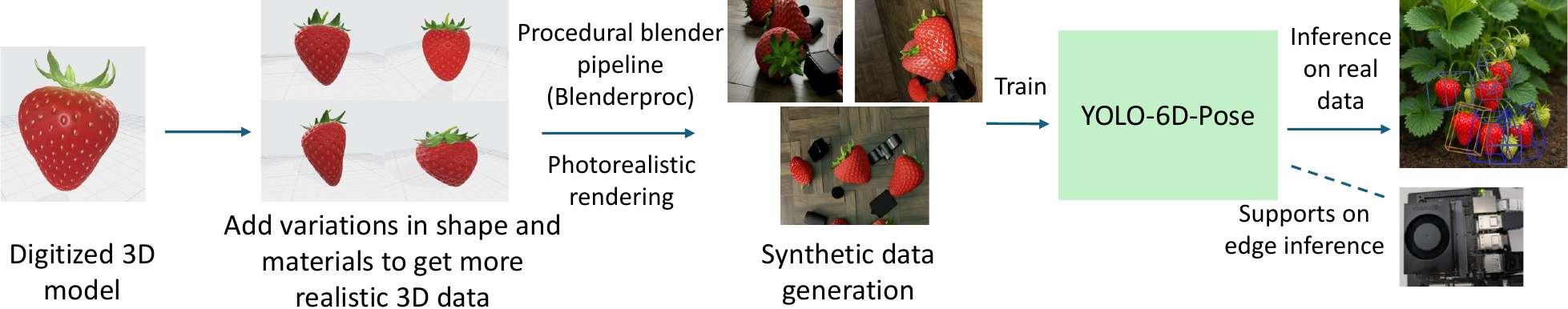}
    \caption{Pipeline for generating synthetic data and then training on it for 6D pose estimation (utilizing the YOLO-6D-Pose~\cite{yolo6d_pose} model) of strawberries. The trained model can also be used for inference on edge devices.}
    \label{fig:pipeline_6d_pose}
\end{figure*}
In the rapidly advancing world of smart farming, the use of robotic systems is revolutionizing fruit harvesting~\cite{duckett2018agriculturalroboticsfuturerobotic}. These cutting-edge technologies enhance production quality by automating various steps during harvesting. With their ability to perform selective color picking, robots can efficiently identify and harvest only the ripest fruits, ensuring that consumers receive fruits of better quality. Operating continuously, these systems boost efficiency and reduce reliance on seasonal labor, leading to significant cost savings for agricultural operations. Their agile design and user-friendly interfaces make them easy to operate, allowing farmers to integrate them smoothly into existing workflows. Hence, the selective harvesting of fruits through robotic technology offers a promising solution to the societal and economic issues related to agricultural labor shortages especially in advanced economies, paving the way for a more sustainable future in farming.

Strawberries rank among the most popular fruits globally, with the strawberry industry's annual retail value surpassing \$17 billion~\cite{parsa2023autonomousstrawberrypickingrobotic}. However, the economic sustainability of this sector is jeopardized by substantial labor costs, which exceed \$1 billion~\cite{parsa2023autonomousstrawberrypickingrobotic} dedicated solely to the selective harvesting process each year. Thereby, strawberries offer significant commercial value in the agricultural sector. However, their harvesting remains a labor-intensive process, with labor costs accounting for approximately 40\% of total production expenses~\cite{li2024singleshot6dofpose3d}. The reliance on seasonal labor in strawberry harvesting leads to increased costs and challenges, especially during peak seasons when labor shortages are common. The COVID-19 pandemic has further highlighted the urgent need for automated solutions that can improve efficiency and reduce costs in the strawberry market. Hence, there is a need for automated harvesting of strawberries. Companies like Organifarms~\cite{organifarms} and Tevel~\cite{tevel} are actively developing automated technologies for picking various fruits, including strawberries, plums, apricots, etc. Most existing robotic vision methods for strawberry picking utilize a 2D to 3D transformation approach~\cite{organifarms, tevel, montoya-cavero2022}. These methods typically employ either traditional image processing algorithms or machine learning techniques to first determine the 2D coordinates of strawberries in an image. This data is then correlated with depth information from specialized depth sensors to create approximate 3D coordinates. However, despite advancements in designing fault-tolerant end-effectors, the overall performance of these robotic systems remains suboptimal due to the incomplete 3D pose information of the target strawberries. A comprehensive understanding of the 6 degrees of freedom (6DoF) pose of each strawberry is crucial for a robotic arm to effectively and safely separate the target fruit especially in a highly cluttered environment. The detection and picking of strawberries is challenging due to their varying shapes and colors. \finalrevision{}{Additionally, the assessment of their
ripeness can be inaccurate and many
ripe strawberries may not become correctly detected as ripe fruits.} This variability complicates the harvesting process, making it difficult for robotic systems to effectively identify and select the optimal fruit. Synthetic data generation can help us simulate these environments for effective pose estimation. 

A recent approach using synthetic data generation~\cite{li2024singleshot6dofpose3d} enables the detection of poses through simulation software that explicitly models strawberry scenes. \finalrevision{}{However, the dataset generation process does not simulate different backgrounds or distractors in a scene limiting robustness in real‐world, heterogeneous settings. It also omits any exploration of edge‐AI optimizations and evaluations—crucial for resource-constrained robotic platforms.} Inspired by this approach, we utilize 3D strawberry models with various shapes and colors, placing them in the scene using BlenderProc \cite{blenderProc}, a procedural pipeline that is free and accessible to the community. This choice \finalrevision{}{enhances our ability} to simulate realistic strawberry \finalrevision{}{appearance}.
\finalrevision{}{Furthermore} \finalrevision{}{building on physics-based} body simulation, we can accurately position the objects within the scene, allowing for more effective training of our model, \finalrevision{}{while also introducing scene complexity in terms of adding occluding distractor objects.} Additionally, we integrate the recent YOLOX 6D pose single-shot~\cite{yolo6d_pose} \finalrevision{}{model}, which improves the detection capabilities for our specific application. Furthermore, we support inference on edge devices, specifically the Jetson Orin Nano, and rigorously evaluate our results on these platforms. This aspect sets our approach apart, as to the best of our knowledge no previous studies utilizing synthetic data for pose estimation of strawberries have addressed the performance of their models on edge devices. By demonstrating the feasibility of running our model on such hardware, we enhance the practicality and accessibility of our solution in real-world agricultural settings. This focus on edge device compatibility helps in developing efficient and deployable technologies for strawberry harvesting on robotic system which often have limited GPU resources.  In light of these advancements, the main contributions of this paper are as follows:
\begin{itemize}
    \item We present a collection of digitized 3D models of strawberries that are specifically designed (various colors and shapes) for synthetic data generation. These models can serve as a foundational resource for developing more accurate and realistic datasets for training pose estimation algorithms.
    \item We introduce a procedural pipeline utilizing BlenderProc~\cite{blenderProc} to generate synthetic data based solely on various strawberry models. This approach not only facilitates the creation of diverse datasets but also allows for easy extensions, such as incorporating different types of distractors and varying environmental conditions to enhance the realism of the simulations. \finalrevision{}{We also release the data generation tool as well as the respective dataset used for this study with our paper.}
    \item Our methodology includes support for inference on edge devices, specifically the Jetson Orin Nano. We provide a thorough evaluation of the model's performance on these platforms, demonstrating its practicality and efficiency for real-time applications in agricultural settings. Inference on edge devices enables deployment across a variety of robotic systems, which typically operate with limited computational resources.
\end{itemize}
\section{Related works}
Monocular 6D pose estimation can be categorized into two primary approaches. One approach involves directly regressing the final 6D pose, while the other relies on establishing 2D-3D correspondences using techniques such as \finalrevision{}on{}{the} Random Sample Consensus (RANSAC) based Perspective-n-Point (PnP) algorithm. Both methodologies can incorporate refinement techniques to enhance the accuracy of the initially estimated pose.
\noindent\textbf{Indirect pose estimation approaches:} The most widely adopted approach for 6D pose estimation involves establishing 2D-3D correspondences prior to applying the RANSAC-based PnP algorithm to solve for the pose~\cite{direct1, direct2, direct3, direct4}. \finalrevision{}{Initial} methods~\cite{direct1, direct2} focused on computing 2D projections of the corners of a 3D bounding box, which serves as a foundational step in the pose estimation process. Subsequently, Peng et al.~\cite{direct3} highlighted an important aspect of pose estimation by demonstrating that utilizing keypoints positioned away from the object's surface can lead to significant errors in the pose estimation results. To address this issue, they proposed a method that samples multiple keypoints directly on the object model, thereby improving the accuracy of the estimates.
To further enhance the robustness of the pose estimation process segmentation techniques combined with a voting mechanism for each correspondence~\cite{direct3, direct4}. This approach helps ensure that the estimated pose is less sensitive to noise and inaccuracies in the detected keypoints.
\noindent\textbf{Direct pose estimation approaches:} Indirect pose estimation approaches cannot be applied for many tasks which require the pose estimation to be differentiable~\cite{indirect_differentiable} and this issue is addressed by employing direct pose estimation approaches. Direct pose estimation methods focus on directly regressing a representation of the 6D pose. \finalrevision{}{The} single Shot Detector (SSD) framework~\cite{indirect_liu} was extended by Kehl et. al.~\cite{indirect_kehl} to estimate the 6D object pose by discretizations of the pose space and employing a classification approach instead of traditional regression methods.
PoseCNN~\cite{xiang2018posecnn} exemplifies a straightforward approach to 6D pose estimation using a convolutional neural network (CNN). It takes a different approach by regressing depth information, the projected 2D center, and a quaternion for each region of interest within a custom detection pipeline. This method incorporates a Hough voting layer to accurately localize the object's center within the image.
DeepIM~\cite{li2018deepim} introduced an innovative iterative refinement technique that regresses the difference between the pose hypothesis rendered image and the actual input image. This approach aims to improve the accuracy of pose estimation by continually adjusting the pose based on the discrepancies observed.
CosyPose~\cite{labbe2020cosypose} further enhances the DeepIM framework by incorporating a continuous rotation parameterization and leveraging more modern neural network architectures, thereby improving the robustness and accuracy of pose estimation.
EfficientPose~\cite{bukschat2020efficientpose} proposed an enhancement to the EfficientDet~\cite{tan2020efficientdet} object detection model to facilitate pose estimation. GDR-Net~\cite{gdr_net} further enhances 6D pose estimation by addressing the limitations of both indirect and direct regression methods, ultimately proposing the Geometry-guided Direct Regression Network (GDR-Net) for improved performance. However, this method faces challenges due to the lack of proper parameterization for the pose parameters. Additionally, the reliance on the vanilla ADD(S) loss function, combined with aggressive scale augmentation techniques, can lead to instability during training, particularly when applied to large datasets such as YCB-V~\cite{xiang2018posecnn}. This challenge is effectively addressed by the YOLO-6D-Pose~\cite{yolo6d_pose}, which not only improves robustness but also supports inference on edge devices through the use of the small or tiny backbone of YOLOX. Consequently, we have chosen this method for \finalrevision{}{estimating} the 6D pose of strawberries.
\noindent\textbf{Pose estimation for agricultural robotics:} In the realm of agricultural robotics, particularly in fruit picking, various 2D detection methods have been deployed effectively. For instance, the YOLO-v4 model has been applied to detect cherry fruits~\cite{cherry}, successfully addressing challenges posed by environmental factors such as shadows and achieving superior performance compared to the standard YOLO-v4 model. Additionally, the YOLO-v4 model has proven effective for detecting oranges~\cite{orange} and multiple fruit types simultaneously~\cite{multiple_fruits}.
Recently, advancements have been made with the introduction of YOLOv9 and YOLOv10 models for real-time detection of strawberry stalks~\cite{strawberry_detect}. However, all of these methods rely on 2D detection techniques, which often necessitate a 3D transformation scheme.
A more recent approach has employed synthetic data alongside a YOLO-based architecture for the pose estimation of strawberries~\cite{li2024singleshot6dofpose3d}. \finalrevision{}{However, this method is focused on strawberries positioned within a specific scene and the provided data does not contain 3D data and respective annotations.} To effectively simulate diverse real-world environments, it is essential to utilize physics-based body simulations in conjunction with various distractors, backgrounds, and scenes.  To overcome these challenges, we employ different strawberry models along with a procedural rendering pipeline from BlenderProc. This pipeline not only facilitates the creation of realistic simulations but can also be utilized by the community for further development. Additionally, we plan to release our \finalrevision{}{models and the dataset used for this study to} to contribute to ongoing research and advancements in this field.
\section{Methodology}
\begin{figure}[htb!]
    \centering
    \includegraphics[width=\linewidth]{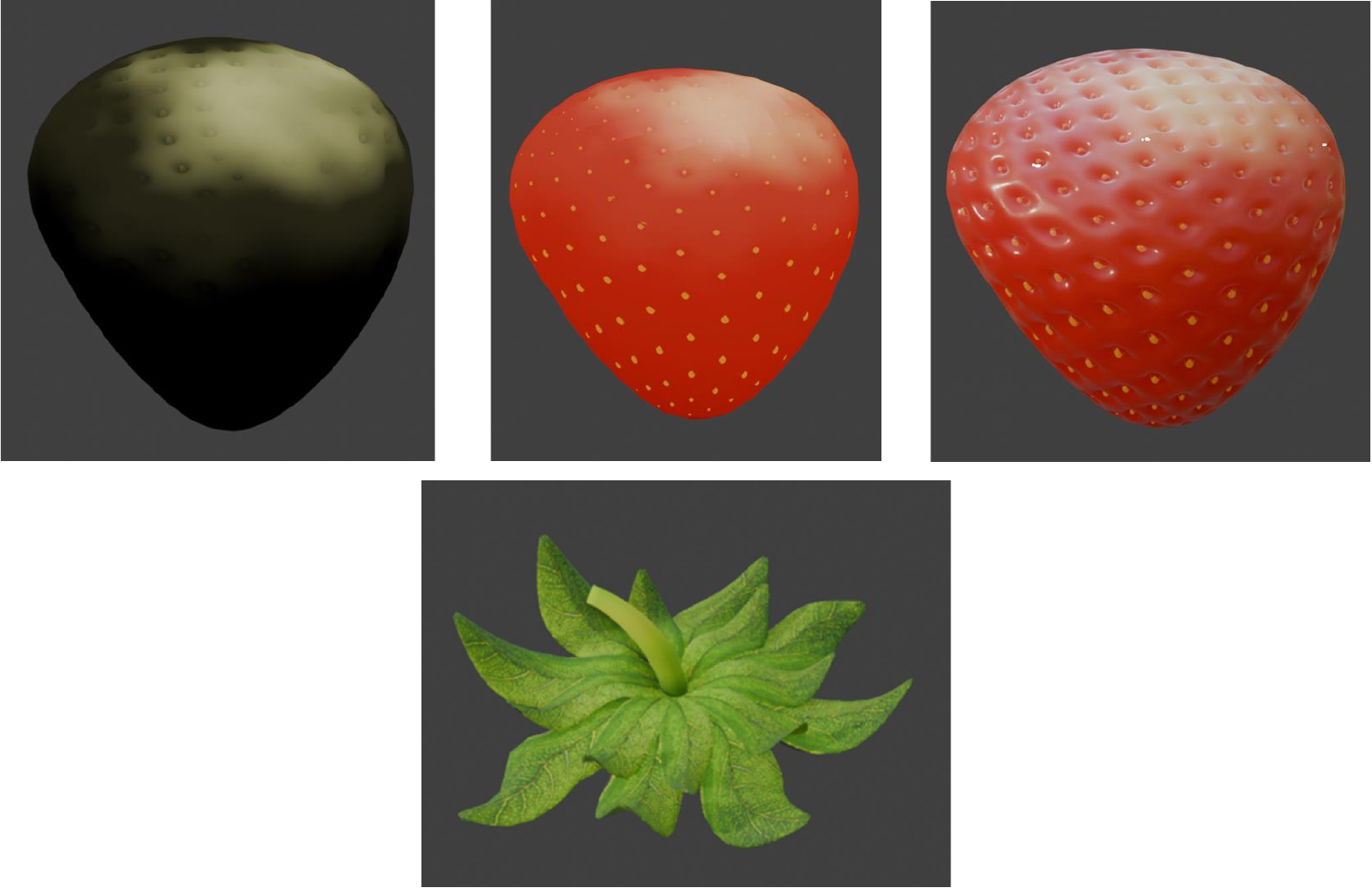}
    \caption{Blending the noise texture with a gradient texture along with slight tint of yellow enhances the strawberry's color by emphasizing variations at the top (left) and transitioning from red to a subtle yellow hue (middle). Finally, some subsurface scattering is applied to get the final output (right). The leaves were modified and a stem was also added (bottom).}
    \label{fig:blending_strawberry_model}
\end{figure}
In the following sections, we will detail the methodology employed for synthetic data generation and pose estimation. This includes an overview of the techniques used to create realistic strawberry models and the implementation of the YOLO-6D-Pose network~\cite{yolo6d_pose} for accurate pose estimation from RGB images.
\subsection{Synthetic data generation}
\begin{figure}[htb!]
    \centering
    \includegraphics[width=\linewidth]{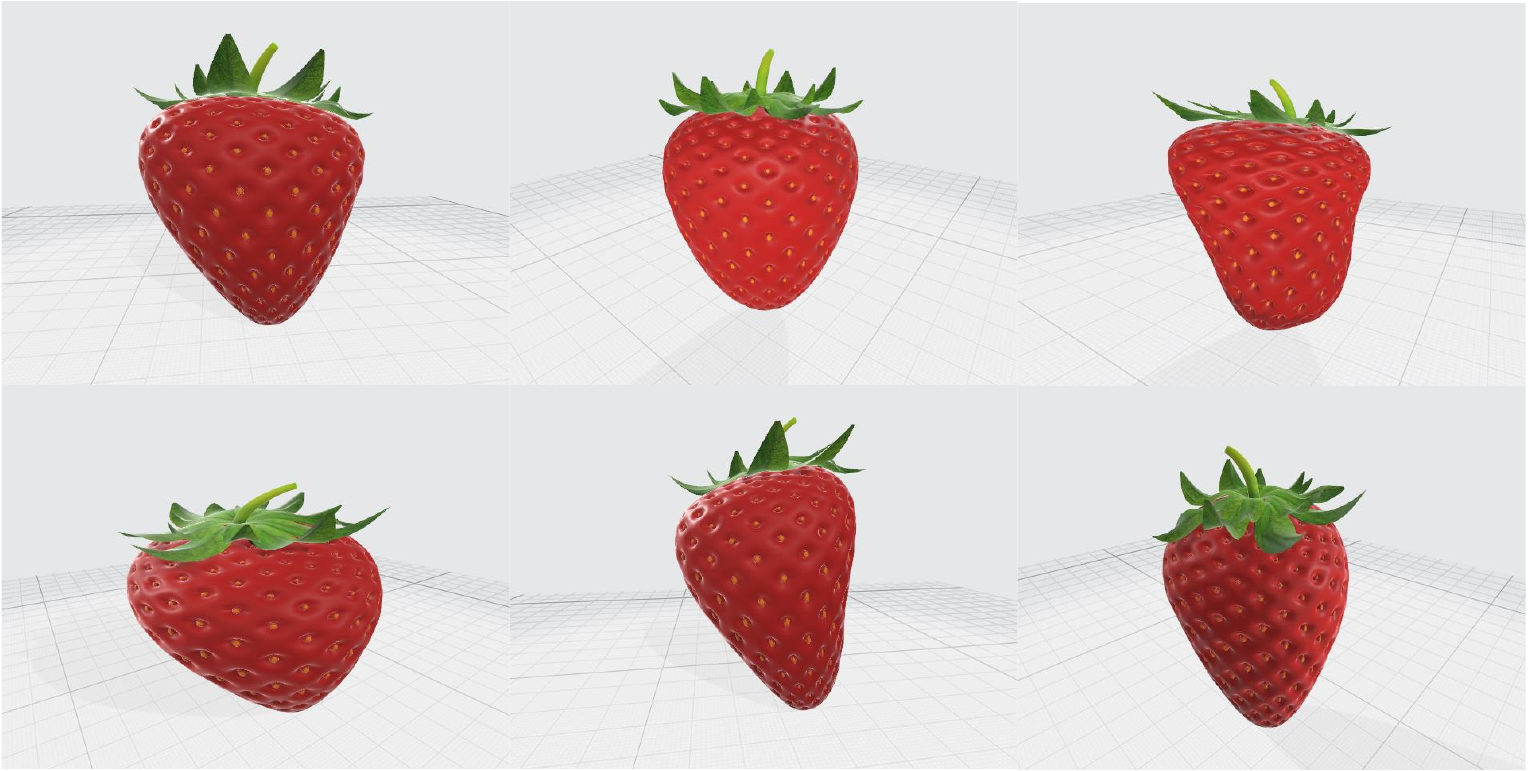}
    \caption{Variations in different shapes of the strawberries used for creating the synthetic data}
    \label{fig:variations_strawberries}
\end{figure}

\begin{figure*}[htb!]
    \centering
    \includegraphics[width=0.8\linewidth]{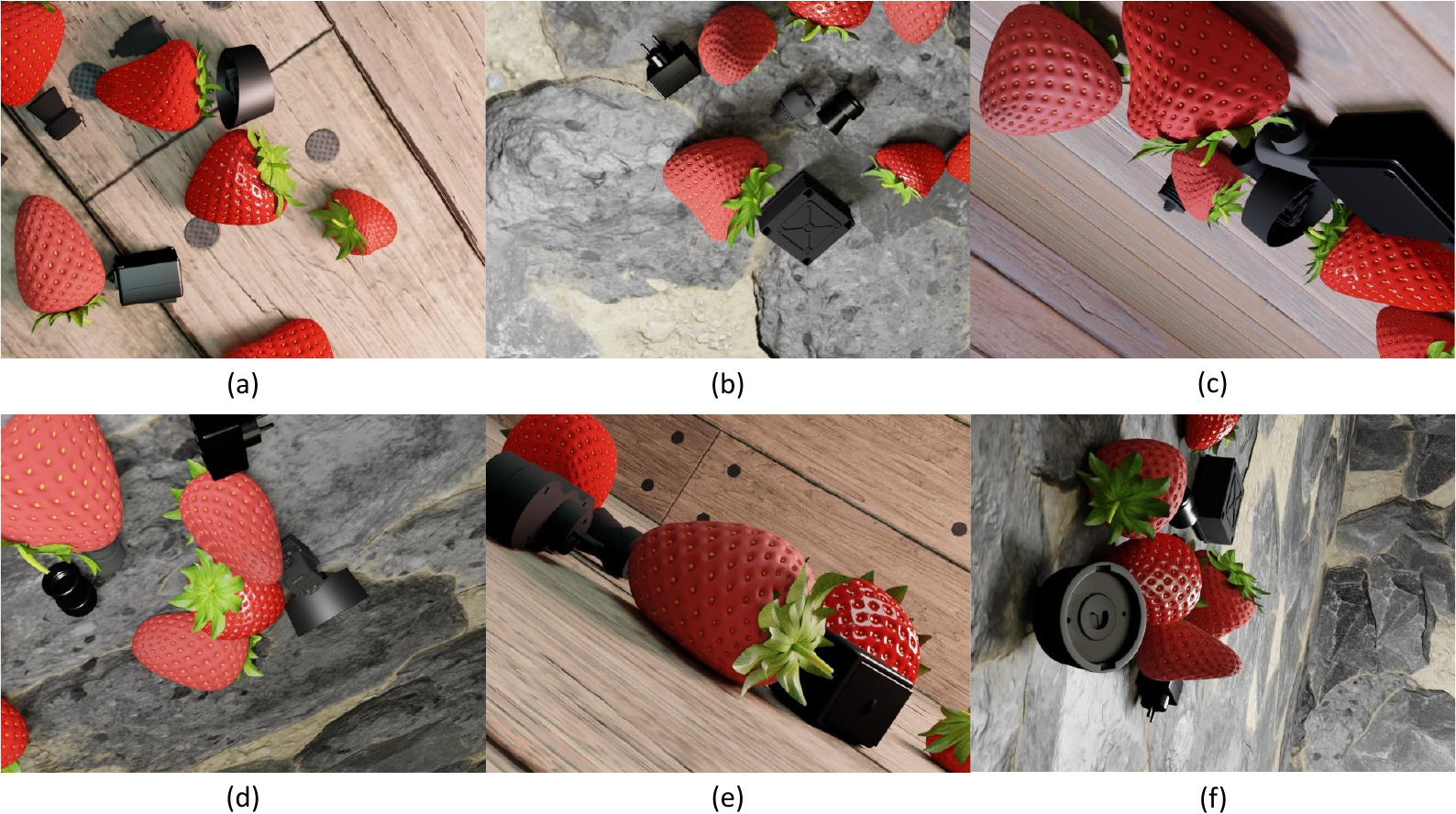}
    \caption{Examples of the synthetic dataset generated using Blender's procedural pipeline~\cite{blenderProc}. The images illustrate variations in materials and lighting conditions (a and b), along with occlusion effects (c). Additionally, the dataset features occlusions created by distractors (d, e, and f) and showcases diverse backgrounds (e and f).}
    \label{fig:strawberries_synthetic_data}
\end{figure*}

In this section, we describe our workflow for synthetic data generation utilizing 3D digitization techniques and simulation using procedural pipelines.
\noindent\textbf{3D digitization:}
We used a base model created by a designer and freely available under a CC-BY license from \finalrevision{}{Sketch-fab}~\cite{gelmi2024strawberry}. In our approach, we initiated various modifications to make the rendering of the strawberries more realistic to bridge the \finalrevision{}{reality gap} with real strawberries \finalrevision{}{and allow a better generalization to data captured in the wild. First,} we utilized Blender’s built-in \finalrevision{}{procedural Perlin Noise texture} to introduce a less uniform appearance to the red \finalrevision{}{colour} of the \finalrevision{}{strawberry’s} surface. This effect was achieved by \finalrevision{}{first subtracting} the Noise texture \finalrevision{}{from} a Gradient Texture. The blending was designed such that the noise is more pronounced at the upper portion of the strawberry, simulating the natural variations found in real fruit. \finalrevision{}{Finally,} this modified texture was employed to blend the vibrant red hue with a subtle yellow (see Figure~\ref{fig:blending_strawberry_model}) tint to have more variations in the hue \finalrevision{}{using the “add” blend mode.}
\finalrevision{}{Furthermore, we also} modified the leaves to \finalrevision{}{better reflect the variance encountered for real-world strawberries} by \finalrevision{}{manually changing the mesh} giving them more random directions. \finalrevision{}{We did this by using Blenders proportional editing, which allows one to edit meshes or polygons so that nearby ones are also affected.  This effect tapers off with the distance to the original selections which makes our leaves more natural.} Additionally, we added a stem, which is just a simple cylinder with a natural curve and some added irregularities (see Figure~\ref{fig:blending_strawberry_model}). It uses the same texture as the leaves. To create different strawberries with variations in shape, we used proportional editing \finalrevision{}{again} to slightly change the shape of each strawberry (see Figure~\ref{fig:variations_strawberries}).
\noindent\textbf{Simulation for dataset generation:}
The six strawberry models (see Figure~\ref{fig:variations_strawberries}) were utilized in a simulation process using Blender's procedural pipeline, known as BlenderProc~\cite{blenderProc}. This method enabled the generation of realistic synthetic strawberry data under a variety of lighting conditions and material settings. The final output consisted of a comprehensive pose estimation dataset, which included RGB images, depth maps, segmentation masks, and annotations formatted according to the BOP standard~\cite{hodan2018bopbenchmark6dobject}.
To enhance realism, \finalrevision{}{we built our simulation on physically-based} rendering (PBR) techniques, which involved modifying the materials, colors, and lighting of the objects within the scene (see Figure~\ref{fig:strawberries_synthetic_data}). Each strawberry was assigned unique textures and colors to \finalrevision{}{capture the variety as encountered in real-world environments}. Cameras were strategically positioned around the scene, sampling multiple viewpoints that focused on the strawberries while maintaining a safe distance to avoid being thrown out of the scene. Additionally, the use of a physics simulation allowed both the strawberries and distractor objects to settle realistically on a surface, mimicking their natural interactions (see Figure~\ref{fig:strawberries_synthetic_data}).
In this context, distractors refer to additional 3D objects included in the scene to create a more complex and realistic environment. These objects help simulate a cluttered setting, which is commonly encountered in real-world scenarios, and they serve to model occlusion effects  (see Figure~\ref{fig:strawberries_synthetic_data}e). By incorporating distractors, the generated dataset becomes more challenging and diverse, thereby enhancing the robustness of learning-based approaches trained on it. For this dataset, distractors were sourced from the YCB-Video (YCBV) dataset~\cite{xiang2018posecnn}, which contains a wide variety of 3D models. \finalrevision{}{By incorporating a wide range of distractor models—not just leaves—we force the detector to learn more discriminative features, which improves its generalizability and leads to more reliable strawberry detection across diverse, real-world scenes.} 
Diverse backgrounds were also simulated by utilizing various textures from the BlenderProc utility~\cite{blenderProc} known as CCTextures (see Figure~\ref{fig:strawberries_synthetic_data}e and Figure~\ref{fig:strawberries_synthetic_data}f). This resource provided a wide range of high-quality, realistic textures that were applied to the environment, enhancing the overall visual diversity of the scenes. By incorporating different backgrounds, the synthetic dataset better mimicked real-world scenarios, where objects are often placed against a variety of surfaces and settings.
\subsection{Pose estimation}
The objective of our approach is to estimate the 6D pose \( P = [R|t] \) for each object \( O \) in an RGB image \( I \), where \( R \) represents 3D rotation and \( t \) denotes 3D translation. We utilize the YOLO-6D-Pose network~\cite{yolo6d_pose} to perform end-to-end pose-estimation as it supports edge inference, making it suitable for deployment in resource-constrained environments. YOLOX, in particular, is a premier algorithm that balances both efficiency and accuracy, allowing for real-time object detection and pose estimation. This makes it an ideal choice for our 6D pose estimation framework, as we aim to detect strawberries in real time using a robotic arm that operates under limited computational resources.
\noindent\textbf{Pose Parameterization:}
YOLO-6D-Pose~\cite{yolo6d_pose} utilizes a 6-dimensional representation \( R_{6d} \) for rotation, defined as the first two columns of the rotation matrix \( R \). This avoids discontinuities present in traditional representations such as quaternions. For translation, it decouples the 3D translation into 2D projected coordinates and depth, ensuring scale and location invariance.
\noindent\textbf{Network Architecture:}
The YOLOX-6D-Pose architecture~\cite{yolo6d_pose} extends the YOLOX~\cite{ge2021yoloxexceedingyoloseries} base model by integrating object detection with pose estimation. The entire network is built upon the CSPDarknet53~\cite{cspnet} backbone coupled with PANet-based feature aggregation~\cite{pathnet}. It consists of a rotation head that predicts the \( R_{6d} \) representation and a translation head $(t)$ that predicts the normalized translation parameters. This unification allows the model to learn both tasks simultaneously, enabling efficient pose estimation from a single forward pass.
\noindent\textbf{Augmentation Techniques:}
A variety of augmentation strategies are employed by the YOLOX-6D-Pose model~\cite{yolo6d_pose}  to improve the robustness and accuracy of \finalrevision{}{the} model. These techniques include end-to-end 6D augmentation, where images are transformed alongside the corresponding poses of all objects. This approach ensures that any modifications applied to the image accurately reflect the changes in the 3D poses of the objects, facilitating better model training.

\finalrevision{}{The YOLOX-6D-Pose architecture~\cite{yolo6d_pose} integrates translation, rotation, and scale augmentations into its data augmentation pipeline, applying the same geometric transformations to both the input image and the 6D poses of all objects present in the image.}
These \finalrevision{}{augmentations} helps maintain alignment between the projected CAD models and the real objects in the image, allowing the model to learn from varied orientations effectively. In addition, the architecture also utilizes scale augmentation, which rescales the images while proportionally adjusting the translation component of the object poses. This aspect is managed carefully to mitigate potential misalignments caused by occlusion or significant rotation angles. By ensuring that scaling is applied judiciously, we maintain the integrity of the pose predictions.

Furthermore, the approach~\cite{yolo6d_pose} applies translation augmentation, where images are randomly translated horizontally and vertically. This adjustment is coupled with corresponding modifications to the pose parameters, allowing the model to learn how to handle variations in object positioning within the scene. The final augmentation strategy systematically combines various transformations: rotation randomly selected from \( (0, 10) \) degrees, translation randomly selected from \( (0, 10)\% \), and scaling with a factor chosen from \( (0.9, 1.1) \). By refining the scale augmentation based on observed discrepancies, the model remains robust against misalignments. Additionally, random color-space augmentations are applied, which do not require any transformations of pose parameters, further enhancing the diversity of our training dataset.
\noindent\textbf{Loss Function:}
To optimize the pose parameters, a combination of various loss functions \finalrevision{}{is} used by the YOLOX-6D-Pose model~\cite{yolo6d_pose}, focusing on the ADD(-S) metric~\cite{adds}, which couples rotation and translation. This approach not only aims to enhance the overall pose estimation accuracy but also to optimize each component of the pose independently to address the limitations of the ADD(-S) loss.

\noindent\textbf{ADD(-S) Loss:}
The ADD(-S) loss is designed to evaluate the accuracy of the estimated poses based on the distance between projected points of the predicted and ground-truth poses. It is defined as follows:
\begin{equation}
\scalebox{0.78}{
\begin{math}
\begin{aligned}
L_{ADD(-S)} = 
\begin{cases} 
L_{asym} = \frac{1}{m} \sum_{x \in m} \left\| (R_p x + t_p) - (R_g x + t_g) \right\|^2_d \\ 
L_{sym} = \frac{1}{m} \sum_{x_1 \in m} \min_{x_2 \in m} \left\| (R_p x_1 + t_p) - (R_g x_2 + t_g) \right\|^2_d 
\end{cases}
\end{aligned}
\end{math}
}
\label{eq:placeholder_label}
\end{equation}
where, \( R_g \) and \( t_g \) denote the ground-truth pose, while \( R_p \) and \( t_p \) represent the predicted pose. The set \( M \) consists of the object's 3D model points, \( m \) is the number of points in the model, and \( d_m \) is the object diameter. 
While \( L_{asym} \) computes the distance directly between corresponding points, \( L_{sym} \) considers the minimum distance from each point in the predicted set to any point in the ground-truth set. However, \( L_{sym} \) can be relaxed, particularly for symmetric objects.
\noindent\textbf{Translation Loss:}
The network predicts the projection of the translation components \([t_x, t_y]\) in the image space, an additional loss term is used to \finalrevision{}{enforce} the accuracy of these 2D predictions. This is achieved by using a simplified version of the object keypoint similarity (OKS) loss~\cite{maji2022yolo} defined as:
\begin{equation}
L_{OKS} = 1 - \text{OKS} = 1 - \exp\left(-\frac{d^2}{2s^2}\frac{1}{b_k^2}\right)
\end{equation}
where, \( d \) is the Euclidean distance between the predicted and ground-truth centroids, \( s_b \) is the area of the object, and \( k \) is a keypoint-specific weight set empirically to 0.1. For the \( t_z \) component of the loss, we employ the Absolute Relative Difference (ARD) loss, defined as:
\begin{equation}
L_{ARD} = 1 - \frac{t_{zp}}{t_{zg}}
\end{equation}
Here, \( t_{zp} \) and \( t_{zg} \) are the predicted and ground-truth values for \( t_z \), respectively.
\finalrevision{}{\noindent\textbf{Rotation Loss:}
The rotation loss follows the YOLO-6D \cite{yolo6d_pose} approach, first introduced in \cite{6d_for_rot} where 3D rotations are represented using a continuous 6D representation derived from the first two columns of the rotation matrix. We compute the L1 loss between the predicted and ground truth 6D rotation vectors, which provides a smooth, discontinuity-free optimization landscape compared to alternative rotation parameterizations such as Euler angles or quaternions.}
Finally, the overall pose loss is formulated as follows:
\begin{equation}
\scalebox{0.8}{
\begin{math}
L_{pose} = \lambda_{ADD(-S)} L_{ADD(-S)} + \lambda_{rot} L_{rot} + \lambda_{OKS} L_{OKS} + \lambda_{ARD} L_{ARD}
\end{math}
}
\label{eq:loss_function}
\end{equation}
where, \( \lambda_{ADD(-S)} \), \( \lambda_{rot} \), \( \lambda_{OKS} \), and \( \lambda_{ARD} \) are empirically chosen weights for each respective loss term, allowing for balanced optimization across all components of the pose.
\begin{figure*}[htb!]
    \centering
    \includegraphics[width=0.8\linewidth]{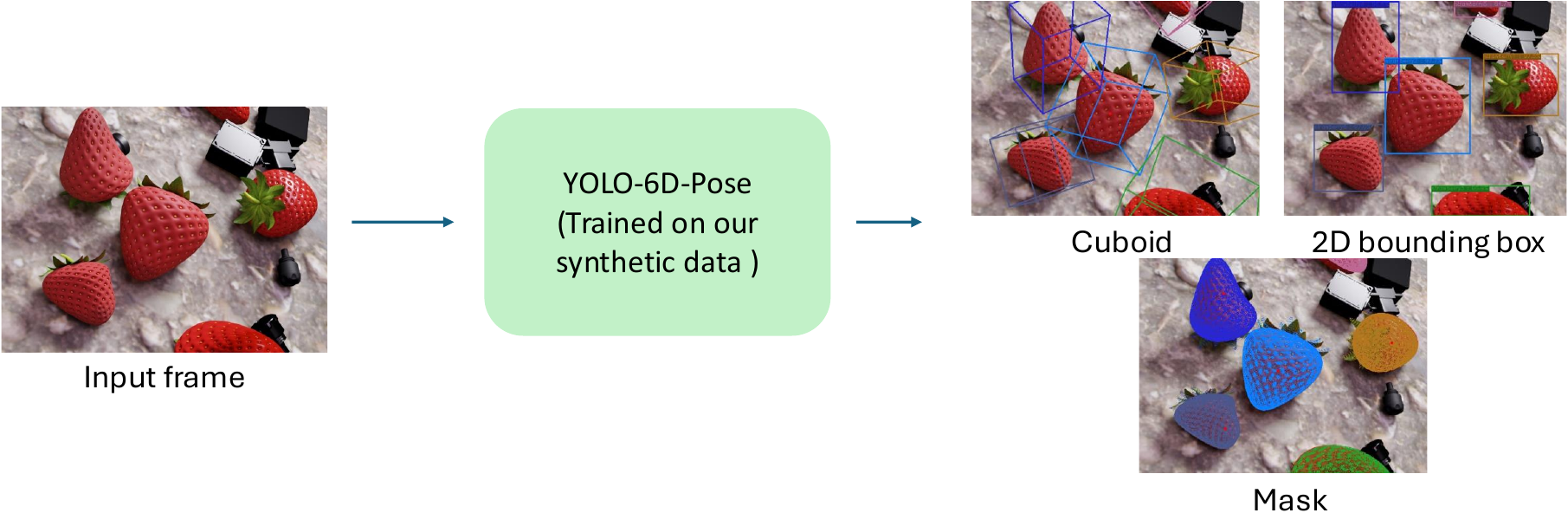}
    \caption{The figure illustrates the output of the YOLO-6D-Pose~\cite{yolo6d_pose} trained using our synthetic data. It outputs the pose, the 2D detections and the corresponding mask.}
    \label{fig:qualitative_output}
\end{figure*}
%

%
\noindent\textbf{Implementation details:} 
The weights \( \lambda_{ADD(-S)} \), \( \lambda_{rot} \), \( \lambda_{OKS} \), and \( \lambda_{ARD} \) (see Equation~\ref{eq:loss_function}) were all set to 1. The training of our model was conducted using four NVIDIA A100 GPUs with a batch size of 32. We employed the Stochastic Gradient Descent (SGD) optimizer, utilizing a cosine learning rate scheduler. The base learning rate was set to \( 1 \times 10^{-2} \), and the model was trained for a total of 300 epochs. We utilized the YOLOX-S backbone for training, as it supports inference on edge devices and all results are presented using this backbone. Additionally, the training process took into account the symmetry of the strawberries.

\section{Evaluation}
In this section, we will provide a detailed evaluation of our 6D pose inference method. We will start with a quantitative assessment, presenting performance metrics such as average processing times for both high-performance and edge devices, along with comparisons of ADD-S metric, rotation and translation errors.
Following this, we will showcase qualitative results from our model applied to synthetic test data and real-world strawberries from a dataset. Through this evaluation, we illustrate both the strengths and areas for improvement in our approach.
\begin{table}[htb!]
    \centering
    \caption{Quantitative analysis of pose inference performance: average forward time, non-maximum suppression (NMS) time, and total average inference time}
    \renewcommand{\arraystretch}{1.5}
    \scalebox{0.85}{
    \begin{tabular}{|l|c|c|}
        \hline
        \textbf{Metric} & \textbf{NVIDIA RTX 3090} & \textbf{Jetson Orin Nano} \\
        \hline \hline
        Average Forward Time               & 22.30 ms         & 35.57 ms         \\
        \hline
        Average NMS Time & 2.21 ms          & 14.02 ms         \\
        \hline
        Average Inference Time             & 24.51 ms         & 49.59 ms         \\
        \hline
    \end{tabular}}
    \label{tab:quantitative_analysis_time}
\end{table}
\subsection{Quantitative evaluation}
\noindent\textbf{Metrics:} We evaluate our method based on the BOP challenge~\cite{hodan2018bopbenchmark6dobject} and the ADD(-S) 0.1d metrics~\cite{adds}. The ADD metric evaluates the average distance of 3D model points between the ground truth pose and the predicted pose, with a pose deemed correct if this average distance is less than 10\% of the object's diameter. In contrast, the ADD-S metric computes the average distance from the predicted pose to the nearest points of the ground truth. For symmetric objects, the ADD-S metric is employed, while for non-symmetric objects, the standard ADD metric is used. Additionally, we compute the rotation error using the formula \emph{rotation error} = $\arccos\left( \frac{\text{trace}(R^T \hat{R}) - 1}{2} \right)$ (in degrees), where \( R \) is the actual rotation and \( \hat{R} \) is the predicted rotation. The translation error is calculated as \emph{translation error} = $||t - \hat{t}||$, where \( t \) is the actual translation and \( \hat{t} \) is the predicted translation. Furthermore, we conduct a performance analysis based on the average non-maximum suppression time, average forward time, and total average time of the network on both the RTX 3090 and the Jetson Orin Nano. This analysis compares edge devices, such as the Jetson Orin Nano, and non-edge devices, like the RTX 3090, highlighting the performance characteristics relevant for various application scenarios.

\noindent\textbf{Performance analysis:} The quantitative analysis of pose inference performance shows that there is \finalrevision{}{no} significant differences between the NVIDIA RTX 3090 and the Jetson Orin Nano (see Table~\ref{tab:quantitative_analysis_time}). The RTX 3090 demonstrates a faster average forward time and non-maximum suppression (NMS) time compared to the Jetson Orin Nano (see Table~\ref{tab:quantitative_analysis_time}). This difference is expected, as the Jetson Orin Nano, being an edge device, has lower computational performance than the high-performance RTX 3090. Additionally, the Jetson Orin Nano is much less costly, making it a more budget-friendly option for certain applications. However, its processing times are still acceptable for use in edge devices, which is crucial for robotic applications.

\begin{table}[htb!]
    \centering
    \caption{Comparison of ADD-S metrics, rotation error, and translation error between high-performance and edge devices}
    \renewcommand{\arraystretch}{1.5} 
    \scalebox{0.8}{
    \begin{tabular}{|l|c|c|}
        \hline
        \textbf{Metric}                        & \textbf{NVIDIA RTX 3090} & \textbf{Jetson Orin Nano} \\
        \hline
        ADD-S\_0p1\_avg                         & 0.7228                    & 0.7231                    \\
        ADD-S\_0p2\_avg                         & 0.7599                    & 0.7594                    \\
        ADD-S\_0p3\_avg                         & 0.7716                    & 0.7716                    \\
        ADD-S\_0p4\_avg                         & 0.7791                    & 0.7791                    \\
        ADD-S\_0p5\_avg                         & 0.7831                    & 0.7832                    \\
        \hline
        rotation\_error\_avg (in degrees)                        & 17.31°                    & 17.70°                    \\
        \hline
        translation\_error\_avg (in mm)                       & 23.10                     & 23.15                    \\
        \hline
    \end{tabular}}
    \label{tab:add_rotation_translation_errors}
\end{table}
\begin{figure*}[htb!]
    \centering
    \includegraphics[width=0.9\linewidth]{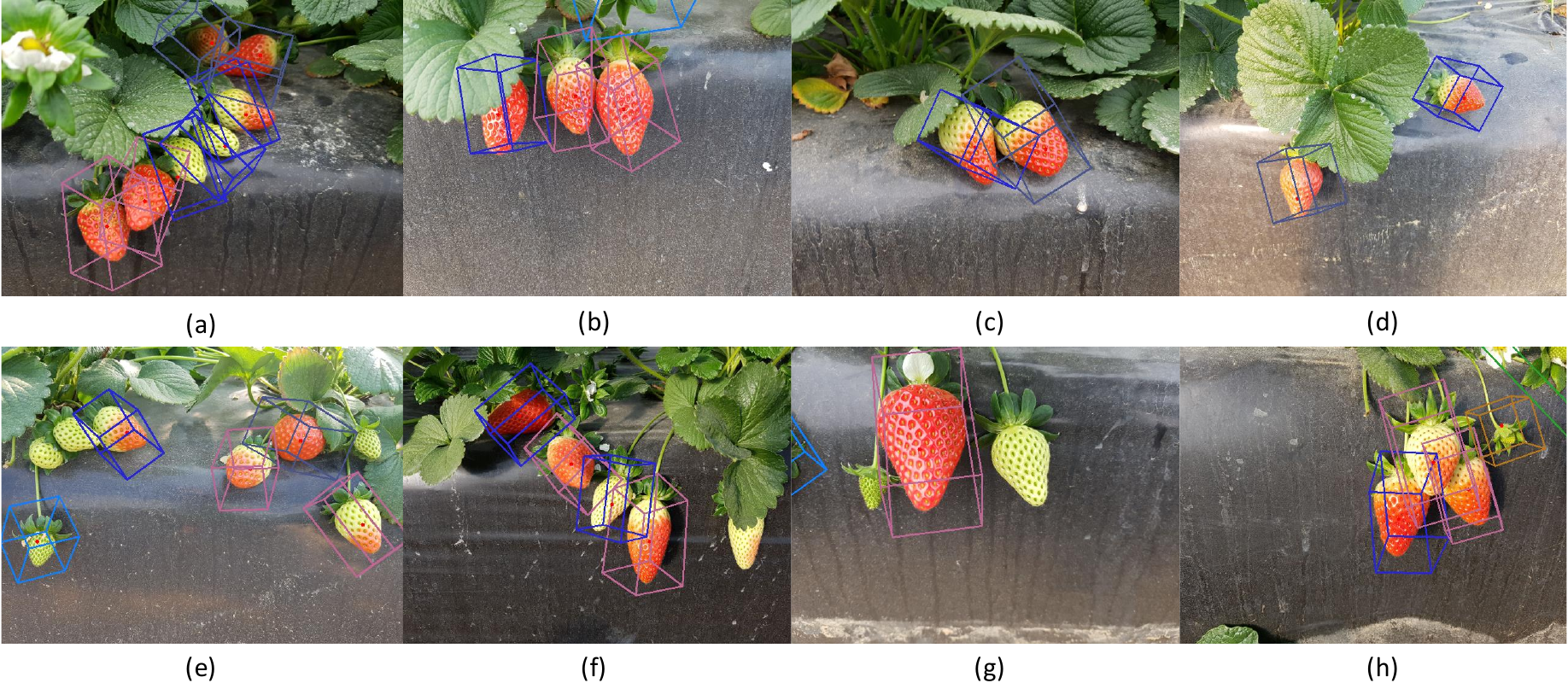}
    \caption{Qualitative analysis of 6D pose inference results on real strawberries from the Strawberry Pose Computer Vision Dataset~\cite{strawberry-pose-7jxqn_dataset}. The results indicate effective detection of ripened and partially ripened strawberries (see (b), (d), and (h)). However, detection of completely unripe strawberries is inconsistent, as shown in images (e), (f), and (g).}
    \label{fig:qualitative_real}
\end{figure*}
\noindent\textbf{Accuracy of pose estimation:} In terms of accuracy, both devices perform similarly across various ADD-S metrics~\cite{adds}, which evaluate the precision of 6D object pose estimation. The metrics ADD-S\_0p1, ADD-S\_0p2, ADD-S\_0p3, ADD-S\_0p4, and ADD-S\_0p5 represent the average distances between the estimated and ground truth poses, with thresholds of 0.1, 0.2, 0.3, 0.4, and 0.5 units, respectively. These thresholds indicate the maximum allowable distance for a successful pose estimation. Across these metrics, the results for the RTX 3090 and the Jetson Orin Nano are almost similar in most cases. Specifically, the average rotation error is slightly lower at 17.31° for the RTX 3090 compared to 17.70° for the Jetson Orin Nano. Similarly, the translation error is comparable, with values of 23.10 mm for the RTX 3090 and 23.15 mm for the Jetson Orin Nano.
Overall, these results indicate that while both devices can achieve similar accuracy, the RTX 3090 excels in speed, making it the preferred choice for applications that require rapid and precise pose inference. In contrast, the Jetson Orin Nano is an excellent option for scenarios where inference is needed in resource-limited environments or on low-power devices.
\begin{figure}[htb!]
    \centering
    \includegraphics[width=0.9\linewidth]{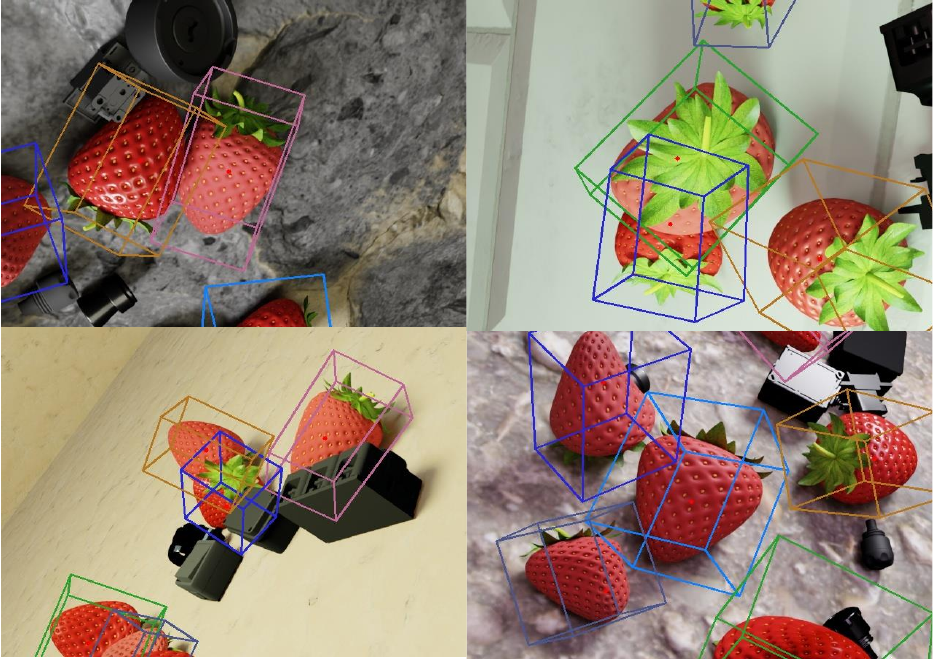}
    \caption{Qualitative analysis of the synthetic dataset, showcasing the accuracy of our model on unseen test data. The results indicate that the model is plausibly capable of inferring the 6D pose of the strawberries. }
    \label{fig:qualitative_synthetic}
\end{figure}

\subsection{Qualitative evaluation}
Figure~\ref{fig:qualitative_output} shows the output of the YOLO-6D-Pose model~\cite{yolo6d_pose} trained on our synthetic data, illustrating its ability to effectively generate pose estimations, 2D detections, and corresponding masks
The qualitative analysis of our model's performance is presented through two datasets: 
our synthetic dataset (only test data that has not been used in training) and a real-world dataset from the Strawberry Pose Computer Vision Dataset~\cite{strawberry-pose-7jxqn_dataset}.

In Figure \ref{fig:qualitative_synthetic}, we showcase the accuracy of our model on unseen synthetic test data. The results demonstrate the model's capability to reliably infer the 6D pose of strawberries, indicating its robust performance in controlled conditions.

Figure \ref{fig:qualitative_real} presents the qualitative analysis of 6D pose inference results on real strawberries. The model effectively detects ripened and partially ripened strawberries, as evidenced in images (b), (d), and (h). However, the detection of completely unripe strawberries proves to be inconsistent, as illustrated in images (e), (f), and (g). This inconsistency highlights the challenges faced by the model when dealing with varying degrees of ripeness, suggesting areas for further improvement in detection accuracy for unripe strawberries.

\section{Conclusion and Future Work}
In this work, we successfully demonstrated the effectiveness of the YOLO-6D-Pose model for 6D pose inference of strawberries using both synthetic and real-world datasets. We created a flexible pipeline and approach for generating synthetic strawberry data from different 3D models using a procedural Blender pipeline. The quantitative evaluation revealed that \finalrevision{}{the NVIDIA RTX 3090 and Jetson Orin Nano achieve virtually identical accuracy} across various ADD-S metrics, with slight variations in rotation and translation errors. Notably, the RTX 3090 significantly outperforms the Jetson Orin Nano in terms of speed, making it the preferred choice for applications requiring rapid pose estimation. However, the Jetson Orin Nano remains an excellent option for use in resource-constrained environments, demonstrating that our model is well-suited for robotic applications across different performance tiers. Overall, our results indicate that the model offers flexibility for future implementations in agricultural settings.

The qualitative analysis further supports our findings, showcasing the model's capability to accurately infer the poses of ripened and partially ripened strawberries while identifying areas for improvement in detecting unripe strawberries. For future work, we plan to explore variations in colors to enhance the detection of unripe strawberries. Additionally, if we intend to exclude unripe strawberries from our analysis, we will incorporate them as distractors to improve the model's robustness. Finally, this approach can also be used for other fruits like apples, peaches, nectarines, apricots, plums, etc.
{
    \small
    \bibliographystyle{ieeenat_fullname}
    \bibliography{main}
}

\end{document}